\documentclass{article}

\usepackage[preprint]{neurips_2026}

\usepackage[utf8]{inputenc}
\usepackage[T1]{fontenc}
\usepackage[czech,english]{babel}
\usepackage[hidelinks]{hyperref}
\usepackage{url}
\usepackage{booktabs}
\usepackage{graphicx}
\usepackage{subcaption}
\usepackage{microtype}
\usepackage{xcolor}
\usepackage{amsmath}
\usepackage{amssymb}
\usepackage{mathtools}
\usepackage{amsthm}
\usepackage{float}
\usepackage[capitalize,noabbrev]{cleveref}

\newcommand{\rev}[1]{#1}
\newcommand{\revsubsection}[1]{\subsection{#1}}
\newcommand{\revsection}[1]{\section{#1}}
\newenvironment{revblock}{}{}
\newcommand{\revfig}[1]{#1}

\theoremstyle{plain}

\theoremstyle{definition}

\theoremstyle{remark}

\title{Platonic Representation Hypothesis on World Models}

\author{%
  \textbf{Wenhow Li}\textsuperscript{*} \quad
  \textbf{Chengwei MA}\textsuperscript{*} \quad
  \textbf{Hui Xiong} \\
  \textbf{Ying-Cong Chen} \quad \textbf{Lei Zhang} \\
  The Hong Kong University of Science and Technology (Guangzhou) \\
  Guangzhou, China \\
  {\normalfont\footnotesize \textsuperscript{*}Equal contribution.}
}

\begin{document}

\maketitle

\begin{abstract}
  World models have demonstrated significant potential for perceiving and simulating complex environments. Despite their strong performance, the fundamental nature of their learned representations remains poorly understood. In this paper, we investigate the \textbf{Platonic Representation Hypothesis} within this domain by proposing the \textbf{Predictive Consistency Assumption}: we posit that the optimization of a shared state transition objective acts as a selective pressure that encourages heterogeneous models to converge toward a shared latent structure. Through systematic experiments with the DINO World Model (DINO-WM), in which we vary visual encoders to create heterogeneous models, we find that capable world models evolve toward \textbf{geometrically similar internal structures}. Moreover, via model stitching, we show that the internal features of one world model can be mapped to another with limited performance degradation, providing evidence of functional compatibility. Our findings suggest that the pursuit of \textbf{predictive consistency} can promote shared, transition-compatible latent structure across world models.

  {\raggedright
  \href{https://sellerbubble.github.io/platonic-representation-hypothesis-on-world-models/}{\textcolor{blue}{\textbf{Project page:} \nolinkurl{https://sellerbubble.github.io/platonic-representation-hypothesis-on-world-models/}}}\par}
\end{abstract}

\section{Introduction}

Recent research in general-purpose artificial intelligence has documented a consistent pattern: as neural networks scale in data volume, model capacity, and task diversity, their internal representation spaces exhibit increasing alignment, converging toward a shared statistical model of reality \citep{Tjandrasuwita2025UnderstandingTE, Morcos2018InsightsOR, Kornblith2019SimilarityON, Bansal2021RevisitingMS}. This observation, formalized as the ``Platonic Representation Hypothesis'' (PRH) \citep{Huh2024ThePR}, proposes that disparate models---even with divergent architectures and training objectives---are developing similar representation spaces that reflect analogous statistical understandings of the underlying world. While significant evidence for this convergence has been documented in static vision and language domains, a fundamental question remains unexplored: Does this Platonic alignment extend to World Models, which must internalize not just static features, but the visual dynamics of our physical world?

\begin{figure}[ht]
  \centering
  \includegraphics[width=0.55\textwidth]{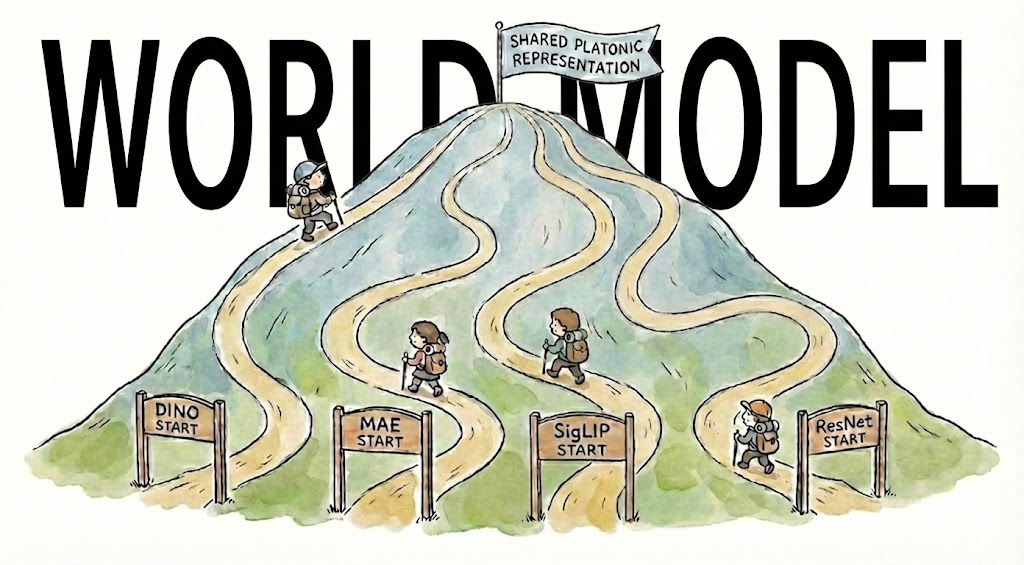}
  \caption{\textbf{The Platonic Representation Hypothesis on World Models.} 
  World models initialized from heterogeneous visual priors are optimized under the same transition objective ($s_t \to s_{t+1}$). Predictive consistency acts as a selective pressure that can drive their internal predictor representations toward a shared latent structure.}
  \label{fig:platonic_hypothesis}
\end{figure}

In this paper, we directly address this scientific gap. World models are trained to simulate the dynamics of a specific environment---which could be a game engine, a robotic simulator, or the real world. We hypothesize that for the same environment being learned, any sufficiently capable world model should converge to a unified Platonic representation. The training objective of a world model is to predict future states of the world based on the current state, following the dynamics that govern the environment. This objective acts as a strong selective pressure, driving diverse world models to adopt shared internal representations, which in turn helps them better capture the underlying laws of the specific world.

To verify this hypothesis, we utilize the DINO World Model (DINO-WM) \citep{Zhou2024DINOWMWM}---a method that models visual dynamics without reconstructing the visual world. We construct a set of ``heterogeneous'' world models by systematically varying the vision encoders \citep{Oquab2023DINOv2LR, He2015DeepRL, Zhai2023SigmoidLF, He2021MaskedAA}, thereby providing the models with diverse visual feature priors. Our analysis focuses on the Predictor---the core component responsible for modeling world dynamics---to determine if its inner representations evolve toward consistency despite receiving varied input signals.

Our experimental results provide evidence for the Platonic Representation Hypothesis in world models while revealing constraints associated with model architecture. First, we show through geometric metrics that predictors initialized with heterogeneous ViT priors---such as DINOv2, SigLIP, and MAE---progressively converge toward a shared latent structure. This alignment is shaped by architectural inductive biases, as evidenced by the persistent topological gap between transformer-based models and the ResNet baseline. Second, we test the functional compatibility of these representations through ``model stitching'' experiments, showing that a simple mapping can preserve planning performance across several pairs of high-quality models.

In conclusion, our work offers an empirical perspective on the essence of world models: predictive training can encourage heterogeneous visual priors to form shared latent transition structure. This convergence provides a concrete setting for studying Platonic-like alignment in embodied prediction systems.

\section{Related Work}

\subsection{Latent Dynamics and World Modeling}

World Modeling has evolved from early recurrent controllers \citep{Ha2018WorldM} to robust latent dynamics frameworks \citep{Hafner2023MasteringDD, Hafner2019DreamTC, Hansen2022TemporalDL} that achieve high performance without domain-specific tuning. Concurrently, a paradigm shift in representation learning has favored abstract state prediction over pixel-level reconstruction, a principle championed by joint-embedding architectures \citep{Assran2023SelfSupervisedLF, Bardes2024RevisitingFP, Grill2020BootstrapYO} to capture essential environmental semantics.

This approach has scaled into generative video foundation models \citep{Liu2024SoraAR, Bruce2024GenieGI, Yang2023LearningIR, Ren2025VideoWorldEK, Zhang2024MoonshotTC, Wang2025WanOA}, which act as general-purpose physical simulators by learning causal interactions from internet-scale data. In this study, we utilize the DINO-WM architecture \citep{Zhou2024DINOWMWM}---which bridges self-supervised visual priors with dynamics learning---as a testbed for representational analysis. The apparent functional convergence across these diverse paradigms prompts our central inquiry: does this shared capability in modeling temporal evolution imply an underlying structural convergence in their latent spaces?

\subsection{The Platonic Representation Hypothesis and Model Alignment}

Our analysis is grounded in the phenomenon of convergent learning, where neural networks with differing initializations often develop functionally similar features \citep{Li2015ConvergentLD, Kornblith2019SimilarityON, Raghu2017SVCCASV, Morcos2018InsightsOR}. This observation has been formalized as the Platonic Representation Hypothesis (PRH) \citep{Huh2024ThePR}, which posits that models converge toward a shared statistical representation of reality as they scale.

To quantify this alignment, researchers utilize topological metrics and functional probes. Relative Representation frameworks \citep{Moschella2022RelativeRE} suggest that while absolute coordinates differ, the internal geometry (e.g., distances) remains consistent---justifying our use of Mutual k-NN as a primary metric. Additionally, Model Stitching \citep{Lenc2014UnderstandingIR} serves as a rigorous test for the functional interchangeability of these latent spaces.

While PRH has been studied in static modalities like vision and language, its extension to action-conditioned state transitions remains largely unexplored. This work bridges that gap by investigating whether the optimization of predictive consistency serves as a generative force driving structural convergence in the latent dynamics of World Models.

\section{Method: Modeling Action-Conditioned State Transitions toward a Unified Latent Representation}

In this section, we formally define the problem of world modeling as a state transition operator and introduce the Predictive Consistency Assumption. We then describe the architectural framework (DINO-WM) used to test this assumption across heterogeneous vision priors. Finally, we detail the analytical protocols---specifically geometric similarity and functional stitching---used to quantify the convergence of latent representations.

\subsection{Problem Formulation: The World Model as a State Transition Operator}

We define a World Model not merely as a perceptual encoder but as a dynamical system that approximates the governing laws of an environment. Consider a discrete-time control process governed by the true transition dynamics of the physical world $\mathcal{W}$, denoted as $s_{t+1} = \mathcal{T}(s_t, a_t)$. In our formulation, the state $s_t$ is constituted by high-dimensional observations $o_t$ (e.g., pixel values). Accordingly, $a_t$ represents the action proposal conditioned on $s_t$, specifying the intended dynamic change required to transition the system from the current state to the next.

The objective of a World Model is to learn a parameterized approximation of this transition. The model consists of two primary components:

\begin{enumerate}
    \item \textbf{A Vision Encoder} $\phi: \mathcal{O} \to \mathcal{Z}_{enc}$, which maps observations $o_t$ to visual feature embeddings $z_t = \phi(o_t)$. These embeddings serve as the input tokens for the predictor.
    \item \textbf{A Predictor (State Transition Operator)} $\psi_\theta$. The predictor processes a history of visual features and actions to predict the future. Crucially, as a deep neural network, the predictor generates a sequence of \textbf{internal latent representations} $h_t^l$ at each layer $l$.
\end{enumerate}

The learning objective is to minimize the divergence between the predictor's output and the encoded future observation:
\begin{equation}
\min_{\theta} \mathbb{E}_{(o, a) \sim \mathcal{D}} \left[ \| \psi_\theta(z_{\le t}, a_{\le t}) - \text{sg}(z_{t+1}) \|^2 \right]
\end{equation}
where $z$ denotes the features extracted by the fixed vision encoder $\phi$, and $\text{sg}(\cdot)$ denotes the stop-gradient operator.

\textbf{The Predictive Consistency Assumption.}
Building upon the Platonic Representation Hypothesis, we propose the \emph{Predictive Consistency Assumption}. Let $\Phi = \{\phi_1, \phi_2, \dots, \phi_n\}$ be a set of distinct vision encoders with differing inductive biases. Let $\Psi = \{\psi_1, \psi_2, \dots, \psi_n\}$ be the corresponding predictors trained to maximize transition accuracy in the same environment $\mathcal{W}$.

We posit that while the input features $z_t$ provided by different encoders may vary significantly, the internal latent representations $h_t$ evolved by the predictors will converge. Specifically, for any pair of predictors $\psi_i, \psi_j$, their learned internal representation spaces $\mathcal{H}_i$ and $\mathcal{H}_j$ will converge toward a geometrically similar structure $\mathcal{H}^*$. 
Formally, we hypothesize that the optimization of state transition prediction acts as a selective pressure for representational alignment.

\subsection{DINO-WM: A Structural Framework for Representation Alignment}

To empirically evaluate the Predictive Consistency Assumption, we adopt the DINO-WM architecture. This framework allows us to isolate the effects of vision priors by systematically varying the encoder $\phi$ while keeping the predictor architecture $\psi$ constant.

\textbf{Heterogeneous Vision Priors.} To introduce controlled variation in the initial feature space, we utilize four distinct fixed encoders: \textbf{DINOv2} \citep{Oquab2023DINOv2LR} (discriminative, object-centric), \textbf{SigLIP} \citep{Zhai2023SigmoidLF}(contrastive, language-aligned), \textbf{MAE} \citep{He2021MaskedAA} (generative, pixel-level reconstruction), and \textbf{ResNet} \cite{He2015DeepRL} (standard convolutional baseline).

\begin{figure}[t]
\centering
\includegraphics[width=\linewidth]{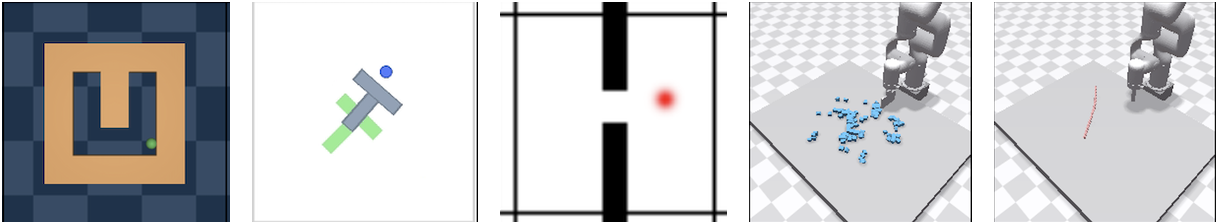}
\caption{\textbf{Evaluation environments.} We evaluate on \textit{PointMaze}, \textit{PushT}, \textit{Wall}, \textit{Granular}, and \textit{Rope}, covering navigation, contact-rich manipulation, and deformable-object dynamics under the DINO-WM benchmark protocol.}
\label{fig:environments_overview}
\end{figure}

\textbf{The Transformer-based Predictor.}
We adopt the predictor architecture directly from DINO-WM \cite{Zhou2024DINOWMWM}. Following their implementation to ensure consistency, we utilize a 6-layer modified decoder-only ViT that operates on patch embeddings without additional tokenization. Our predictor employs a frame-level causal attention mechanism, treating the patch vectors of a single observation as a cohesive unit. It predicts the entire set of next-frame patches $z_t$ simultaneously based on the history of latents $z_{t-H:t-1}$ and actions $a_{t-H:t-1}$, effectively capturing global structural dependencies and temporal dynamics. The detailed settings of the predictor are provided in Appendix~\ref{app:experiment_setting}.

\subsection{Quantifying Convergence: Metrics for Representation Alignment}

We employ two rigorous analytical protocols to measure the convergence of representations: geometric analysis (structure) and model stitching (function).

\textbf{1. Geometric m-kNN Analysis.}
We compare predictor hidden-state geometry using Mutual k-Nearest Neighbors (m-kNN) on paired validation trajectories. The score measures whether two models assign similar local neighborhoods to the same world states; higher values indicate stronger preservation of latent topology. We compute m-kNN per predictor layer and report the layer average. Full definitions are provided in Appendix~\ref{app:metric_stitching_details}.

\textbf{2. Functional Representation Stitching.}
Because geometric alignment alone does not establish planning-usable equivalence, we also splice predictor layers from two trained world models and train only lightweight MLP adapters between frozen components. We evaluate the stitched model with downstream planning success (SSR). High SSR indicates that the two predictors expose exchangeable transition factors rather than merely similar local neighborhoods. Architectural details and directional stitching protocol are given in Appendix~\ref{app:metric_stitching_details}.

\begin{table*}[t]
    \caption{\textbf{Zero-shot Planning Success Rates at Epoch 10.} Comparison of World Models initialized with heterogeneous vision priors across five environments. \textbf{DINOv2-S} is designated as the Platonic Anchor due to its superior average performance and robustness. Success Rates (SR) are reported for navigation/pushing tasks, while Chamfer Distance (CD) is used for Rope/Granular (where lower is better).}
    \label{tab:planning_scores}
    \vskip 0.15in
    \centering
    \begin{small}
    \resizebox{\linewidth}{!}{%
    \begin{tabular}{lccccc}
        \toprule
        Model Prior & PointMaze (SR$\uparrow$) & PushT (SR$\uparrow$) & Wall (SR$\uparrow$) & Rope (CD$\downarrow$) & Granular (CD$\downarrow$) \\
        \midrule
        \multicolumn{6}{l}{\textit{Heterogeneous Priors}} \\
        ResNet         & 0.76 & 0.26 & 0.30 & 1.73 & 0.85  \\
        MAE (ViT-B)    & 0.88 & 0.72 & 0.56 & 1.27 & \textbf{0.30}  \\
        SigLIP (ViT-B) & \textbf{1.00} & 0.68 & 0.86 & 1.33 & 0.42  \\
        \midrule
        \multicolumn{6}{l}{\textit{DINOv2 Scaling Variants}} \\
        \textbf{DINOv2-S (Anchor)} & \textbf{1.00} & \textbf{0.90} & 0.90 & 1.24 & 0.50  \\
        DINOv2-B       & 0.72 & 0.88 & 0.82 & 1.19 & 0.60  \\
        DINOv2-L       & 0.74 & \textbf{0.90} & 0.78 & \textbf{1.11} & 0.48  \\
        DINOv2-G       & 0.74 & 0.88 & \textbf{0.92} & 1.12 & 0.48  \\
        \bottomrule
    \end{tabular}%
    }
    \end{small}
    \vskip -0.1in
\end{table*}

\begin{figure}[t]
  \centering
  \makebox[\textwidth][c]{\includegraphics[width=1.08\textwidth]{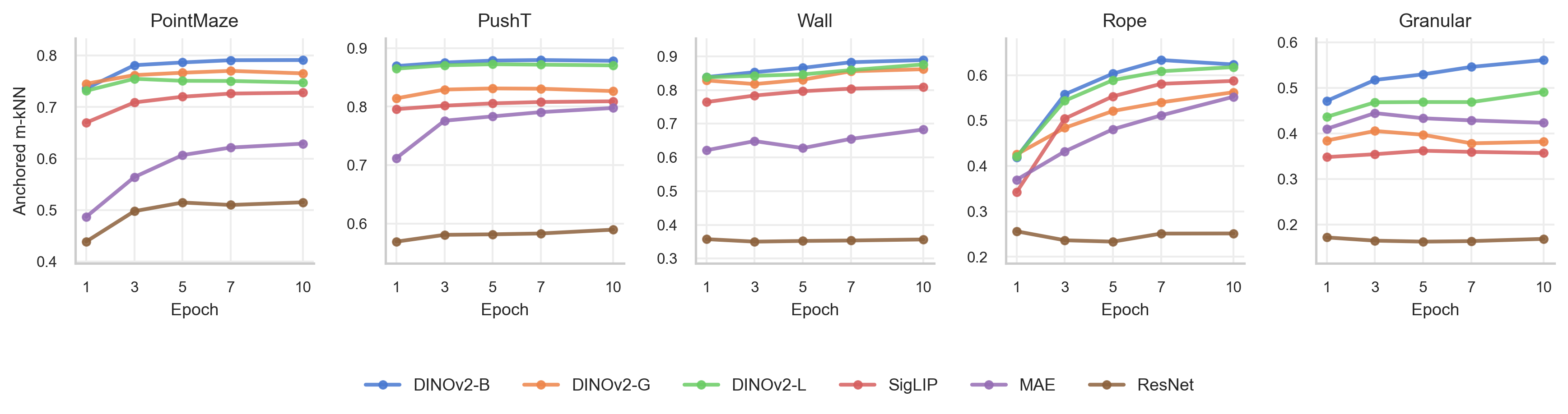}}
  
  \vspace{0.1in} 
  
  \caption{Evolution of Geometric Similarity toward the Platonic Anchor.
  Plots showing the Anchored m-kNN Similarity Score (Y-axis) vs. Training Epoch (X-axis) across five environments. The DINOv2-S World Model (10-epoch checkpoint) serves as the fixed reference Anchor.
  \textbf{Results:} Models initialized with DINOv2-B/G/L (blue/orange/green) exhibit the highest degree of alignment. Meanwhile, MAE (red) and SigLIP (brown) models demonstrate a distinct upward trend in similarity scores across most tasks, indicating progressive convergence. In contrast, the ResNet-based model (pink) consistently yields the lowest scores, suggesting a persistently different representational topology from the ViT-based Anchor in our setting.}
  \label{fig:platonic_evolution}
\end{figure}

\section{Experiments: Evaluating the Platonic Convergence of World Transitions}

In this section, we empirically evaluate the Predictive Consistency Assumption. We first establish the experimental protocol, defining the heterogeneous vision priors and selecting a performance-based anchor. Subsequently, we present quantitative evidence for emergent similarity of geometric structures in the latent space.

\subsection{Experimental Setup: Heterogeneous Priors and the Platonic Anchor}

To test whether models converge despite different visual origins, we train DINO-WM predictors on fixed encoders spanning both modern ViT priors and a convolutional baseline: DINOv2-S/B/L/G, SigLIP, MAE, and ResNet18. The five benchmark environments are \textit{PointMaze}, \textit{Wall}, \textit{PushT}, \textit{Rope}, and \textit{Granular}, with configurations matched to DINO-WM \citep{Zhou2024DINOWMWM}.

\begin{figure}[t]
    \centering
    \includegraphics[width=0.59\columnwidth]{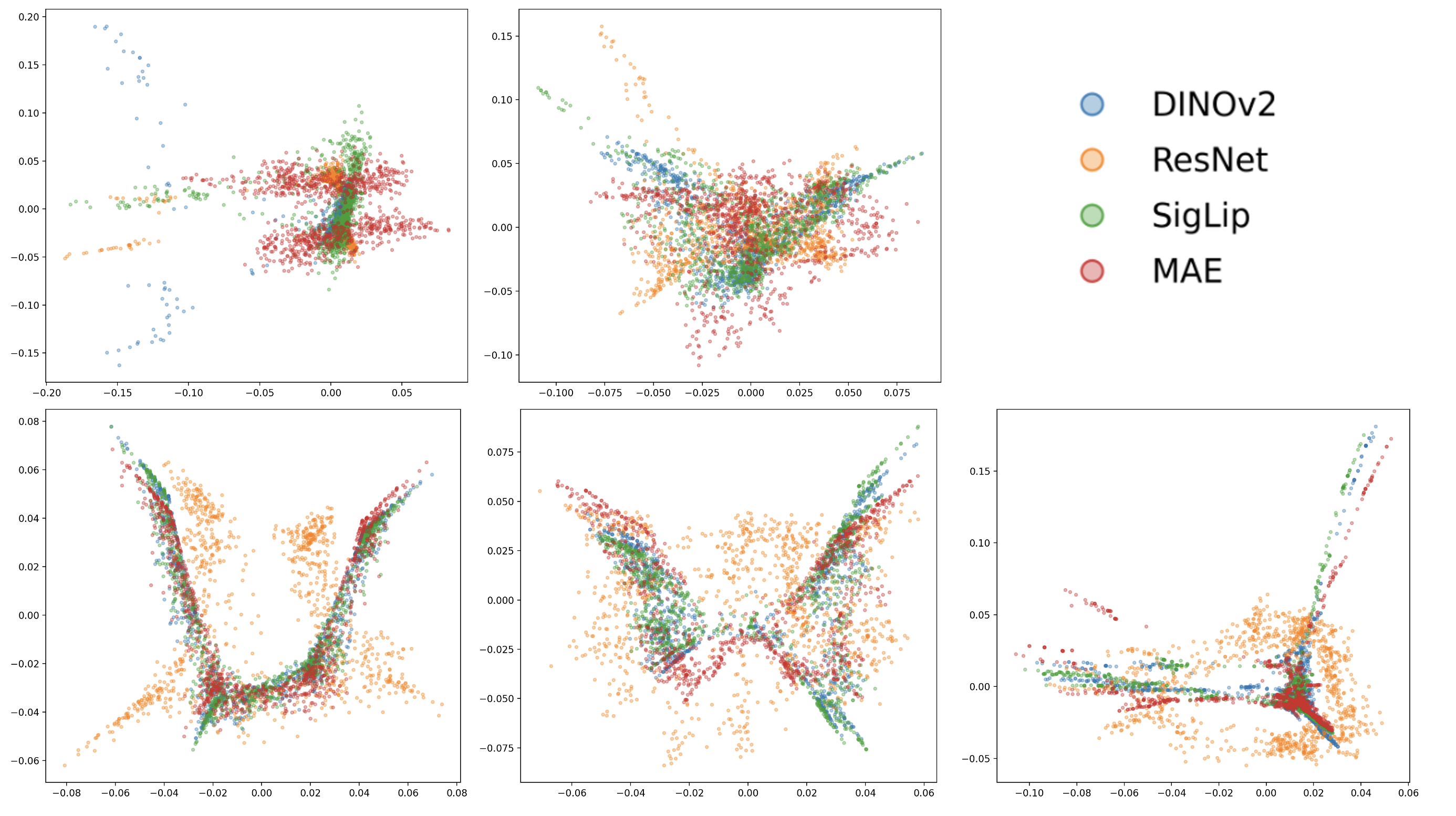}
    \caption{\textbf{Visualizing Representational Alignment via Spectral Embeddings.} 
    We illustrate the geometric structure of predictor hidden states using Laplacian Eigenmaps for five control environments: \textit{Granular}, \textit{Rope}, \textit{PointMaze}, \textit{Wall}, and \textit{Push-T}. 
    Internal representations from heterogeneous vision priors are projected into a low-dimensional space and aligned to the DINOv2 anchor via Procrustes transformation. 
    In the bottom row (\textit{PointMaze}, \textit{Wall}, \textit{Push-T}), a clear pattern of geometric similarity emerges among the transformer-based priors (DINOv2, SigLIP, and MAE), while the ResNet prior (orange) exhibits clear topological misalignment.
    In the top row, \textit{Granular} displays a noisier convergence pattern, whereas the \textit{Rope} environment manifests as a highly concentrated cluster across all models, reflecting the unique dynamical constraints of deformable object manipulation manifolds. Detailed visualization protocols and comprehensive results are provided in Appendix~\ref{app:knn_overlay}. }
    \label{fig:spectral_alignment_single}
\end{figure}

We evaluate zero-shot planning at epoch 10 using the DINO-WM latent-space planner (Appendix~\ref{app:planning_details}). Table~\ref{tab:planning_scores} shows that DINOv2-S provides the strongest and most stable performance among the candidates, so we use its epoch-10 predictor as a performance-motivated anchor for geometric comparison. Candidate models are evaluated at epochs 1, 3, 5, 7, and 10 with anchored m-kNN against this fixed reference. The anchor is not assumed to be a true Platonic ideal; it is an operational reference selected for robust planning performance.

\subsection{Experiment 1: Emergent Structural Similarity in Latent Representations}

\textbf{Objective.} This experiment investigates whether optimization of a shared state-transition objective serves as a selective pressure that encourages world models initialized with heterogeneous vision priors to develop shared latent geometry.

\textbf{Quantitative Analysis.} \Cref{fig:platonic_evolution} illustrates the evolution of Anchored m-kNN Similarity scores across five diverse environments. Figure~\ref{fig:spectral_alignment_single} provides a visualization of the relative relationships between the features of world models initialized with different vision encoder priors. Our analysis reveals distinct geometry dictated by the underlying model architecture:

\begin{itemize} \item \textbf{High-Baseline Alignment (DINO Family):} Predictors initialized with \textbf{DINOv2} scaling variants exhibit the highest overall geometric similarity to the anchor from the early stages of training. While they show a steady upward trend, the alignment phenomenon is relatively less pronounced compared to other ViT priors, as these models already operate within a high-similarity regime consistent with the anchor's architectural family. 

\item \textbf{Clear Progressive Convergence (SigLIP \& MAE):} In contrast, \textbf{SigLIP} and \textbf{MAE} priors demonstrate the largest observed gains in structural alignment. Despite beginning from disparate representational origins, these models exhibit a clear trajectory toward the anchor's latent geometry. This progressive convergence suggests that the pursuit of predictive consistency can reconfigure their heterogeneous feature spaces toward shared, transition-compatible structure.

\item \textbf{Persistent Topological Discontinuity (ResNet):} The \textbf{ResNet}-based model does not exhibit clear convergence toward the ViT-based anchor, with similarity scores remaining near a low baseline. This persistent gap indicates a difference between convolutional and transformer-based inductive biases that is not bridged by the transition objective within the observed training window.

\end{itemize}

\textbf{Insight: Architectural Constraints on Representational Learning.} The experimental results suggest that while optimization for world dynamics can drive representational alignment, \textbf{model architecture imposes a strong constraint} on the learning process in our setting. The convergence observed among the ViT-based models (DINO, SigLIP, MAE) and their collective divergence from the ResNet baseline indicate that architectural compatibility may facilitate recovery of a shared latent representation. Notably, these geometric similarities appear to be partly independent of downstream planning proficiency, suggesting that the evolution of latent structure and the acquisition of task-specific capabilities can follow distinct learning dynamics.

\begin{revblock}
We further validate this structural trend with two robustness checks, reported in Appendix~\ref{app:anchor_sensitivity} and Appendix~\ref{app:extended_training}. First, the qualitative ViT-vs-ResNet alignment pattern remains visible when using SigLIP@10 or MAE@10 as alternative anchors, after removing trivial self-anchor comparisons. Second, the same broad ordering persists at later checkpoints around epochs 20, 30, and 50, although the scores do not increase monotonically with training.
\end{revblock}

\revsubsection{Disentangling Encoder Similarity from Predictor Alignment}

\begin{revblock}
A natural alternative explanation is that predictor-level alignment merely reflects similarity already present in the frozen visual encoders. We test this explanation directly by comparing three quantities on the same validation trajectories: cross-model m-kNN in the raw encoder features at $o_t$, cross-model m-kNN in predictor hidden states at epoch 1, and cross-model m-kNN in predictor hidden states at epoch 10. If the predictor simply transmitted encoder geometry forward, these three measurements should remain close.
\end{revblock}

\begin{figure}[t]
  \centering
  \revfig{\includegraphics[width=0.97\textwidth]{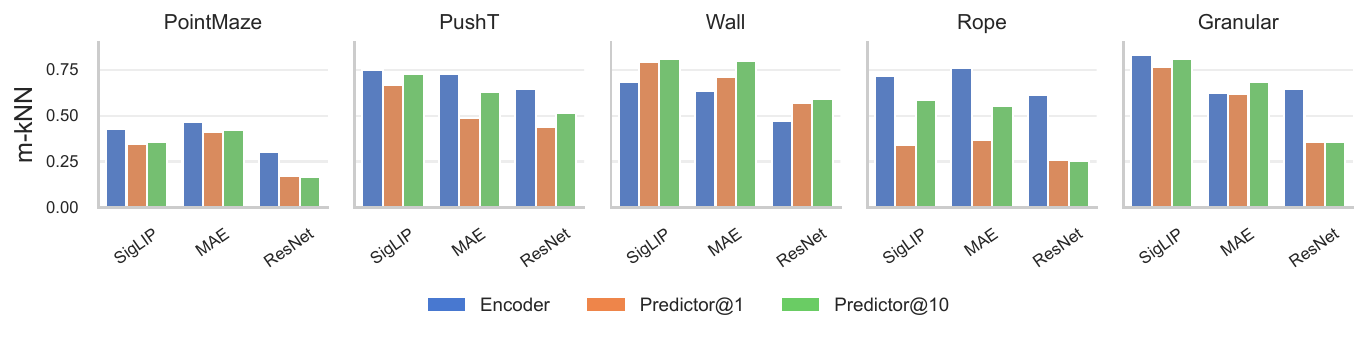}}
  \caption{\rev{\textbf{Encoder-space baseline versus predictor-space alignment.}
  For each environment, we compare m-kNN alignment between DINOv2-S and each candidate model in raw encoder space, predictor space at epoch 1, and predictor space at epoch 10.
  The drop from Encoder to Predictor@1 shows that the predictor does not simply inherit encoder geometry.
  The recovery from Predictor@1 to Predictor@10 then indicates that transition-prediction training induces a new cross-model alignment in predictor space.}}
  \label{fig:encoder_vs_predictor}
\end{figure}

\begin{revblock}
Figure~\ref{fig:encoder_vs_predictor} gives a direct decomposition of the effect. First, frozen encoders already exhibit non-trivial cross-model similarity, so encoder geometry is a real baseline rather than a nuisance to ignore. Second, moving from raw encoder features to epoch-1 predictor states usually changes the score substantially, and most often lowers it. This drop is important: the predictor has already transformed the visual feature space, so the latent alignment measured inside the predictor cannot be explained as a direct copy of encoder-space alignment. Third, from epoch 1 to epoch 10, the predictor-space score often rises again, especially for ViT-based priors. \textbf{Thus, training under a common transition objective does not merely preserve an inherited geometry; it first reparameterizes that geometry and then drives the resulting predictor states toward a more shared latent structure.}
\end{revblock}

\begin{figure}[t]
  \centering
  \revfig{\includegraphics[width=0.92\textwidth]{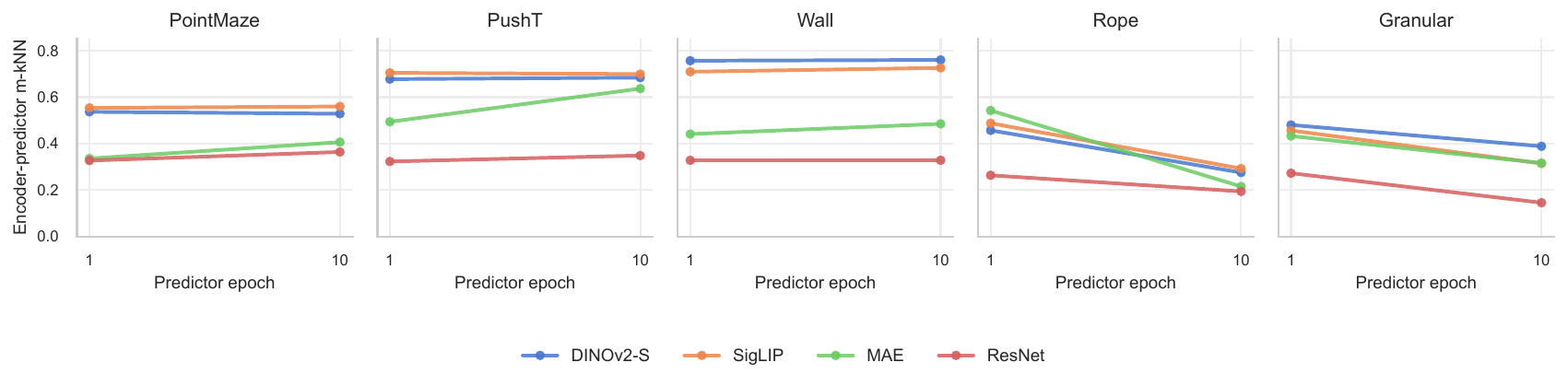}}
  \caption{\rev{\textbf{Within-model encoder--predictor geometry.}
  For each individual model, we compute m-kNN between its own raw encoder features and its predictor hidden states at epochs 1 and 10.
  The predictor retains some encoder information, but the epoch-10 states do not generally become more encoder-like.
  This rules out the interpretation that the cross-model recovery in Figure~\ref{fig:encoder_vs_predictor} is simply a return to raw encoder geometry.}}
  \label{fig:within_model_encoder_predictor}
\end{figure}

\begin{revblock}
Figure~\ref{fig:within_model_encoder_predictor} further separates ``alignment recovery'' from ``encoder recovery.'' If the epoch-10 improvement in Figure~\ref{fig:encoder_vs_predictor} simply reflected predictors drifting back toward their frozen encoders, then within-model encoder--predictor m-kNN should increase in tandem. This is not the dominant pattern. In several settings, such as Rope and Granular, predictors become less encoder-like by epoch 10 even when cross-model predictor alignment improves. \textbf{The two figures therefore support the same conclusion from complementary angles: predictor training changes the inherited encoder geometry, and the later increase in cross-model alignment is better interpreted as emergence of a shared transition-compatible latent structure rather than restoration of the original visual encoder neighborhoods.}
\end{revblock}

\subsection{Experiment 2: Functional Usability via Representation Stitching}

Anchored m-kNN measures preservation of local neighborhood structure across latent spaces, and tests whether task-specific training drives a world model's latent geometry toward a Platonic-style representation. However, geometric alignment does not imply functional universality. To test whether Platonic-like proximity corresponds to planning-usable \emph{shared transition structure}, we adopt \emph{representation stitching}. Specifically, we cut the predictor stacks of two world models at a chosen depth and splice them together, training only lightweight MLP alignment modules to map between feature spaces while freezing all other parameters. The resulting stitched model is evaluated on downstream planning using \textbf{SSR}.

\paragraph{Setup and notation.}
We evaluate on \textit{PointMaze} and \textit{Wall}, where the unstitched world models attain strong planning performance, making stitching a sharper functional test. Our stitching uses three heterogeneous vision priors (DINOv2, SigLIP, MAE) for robustness; ResNet is excluded since its convolutional topology is incompatible with the patch-token predictor interface. Each predictor has 6 layers, and we denote a stitched configuration as $A(1{:}k)+B(k{+}1{:}6)$, with lightweight MLP adapters bridging the feature spaces (Appendix~\ref{app:metric_stitching_details}).

\paragraph{Calibrating the split depth.}
The split depth $k$ can materially affect stitching, since planning-relevant factors may arise at different depths across priors and environments. To avoid an exhaustive sweep over all pairs, we calibrate $k$ on a representative strong transfer, DINOv2$\rightarrow$SigLIP, by scanning $k\in\{1,2,3\}$. The scan shows strong depth sensitivity in \textit{Wall}, with SSR increasing from 0.30/0.38 to 0.62 as more front-end DINO layers are retained, while \textit{PointMaze} exhibits only a modest gain (0.86/0.86 to 0.90). We therefore fix $k=3$ for all remaining pairings to standardize the protocol and limit search, and report the full calibration scan in Table~\ref{tab:ssr_stitching_pointmaze_wall}.

\paragraph{Normalized retention (operational criterion).}
Because raw SSR depends on the absolute competence of both endpoints, it is not directly comparable across pairs. We therefore report a normalized \textbf{retention score(RS)},
\[
\mathrm{Retention}(A\!\rightarrow\!B,k)=\frac{\mathrm{SSR}(A(1{:}k)+B(k{+}1{:}6))}{S_A \cdot S_B},
\]
where $S_A$ and $S_B$ are the unstitched planning success rates of models $A$ and $B$ in the same environment. 

\paragraph{Results.}
Table~\ref{tab:ssr_stitching_pointmaze_wall} summarizes stitching performance on \textit{PointMaze} and \textit{Wall}. We find that when
both endpoint world models are strong planners, which we use as a proxy for proximity to a \textbf{Platonic-like
transition representation}, stitched performance can often be recovered with high RS using only lightweight MLP
alignment. This suggests that high-quality predictors expose exchangeable planning-relevant transition factors that can
be mapped across models while preserving downstream planning.

The effect is strongest in \textit{PointMaze}: under the fixed $k=3$ protocol, many pairings retain high SSR, indicating
substantial shared structure across heterogeneous priors. \textit{Wall} is more discriminative and shows stronger depth
sensitivity. For DINOv2$\rightarrow$SigLIP, increasing the retained front-end depth from $k=1$ to $k=3$ improves SSR
from 0.30/0.38 to 0.62, consistent with a transferable ``Platonic core'' being concentrated in early-to-mid predictor
layers in a harder environment. We defer directionality and reverse stitching analysis to the subsequent experiments.
Many strong-recovery cases achieve \textbf{RS $\ge 0.7$}, which we use as a convenient marker of substantial functional
reuse, without relying on this specific cutoff.

\begin{table}[htbp]
\centering
\caption{Stitched Success Rate (SSR) for different predictor layer combinations.}
\label{tab:ssr_stitching_pointmaze_wall}
\begin{tabular}{p{0.46\columnwidth}cc}
\toprule
\textbf{Layer Combination} & $\mathbf{SSR}_{\text{PointMaze}}$ & $\mathbf{SSR}_{\text{Wall}}$ \\
\midrule
DINO(1) + SigLIP(2--6)        & 0.86 & 0.30 \\
DINO(1--2) + SigLIP(3--6)     & 0.86 & 0.38 \\
DINO(1--3) + SigLIP(4--6)     & 0.90 & 0.62 \\
DINO(1--3) + MAE(4--6)        & 0.84 & 0.42 \\
SigLIP(1--3) + DINO(4--6)     & 0.76 & 0.64 \\
SigLIP(1--3) + MAE(4--6)      & 0.88 & 0.50 \\
MAE(1--3) + DINO(4--6)        & 0.74 & 0.02 \\
MAE(1--3) + SigLIP(4--6)      & 0.72 & 0.00 \\
\bottomrule
\end{tabular}
\end{table}

\paragraph{Directional Stitching.}
We additionally inspect directionality by swapping the front-end/back-end order (A$\rightarrow$B vs.\ B$\rightarrow$A). 
Pairs of strong planners (e.g., DINOv2 and SigLIP) tend to exhibit near-symmetric recovery, supporting functional interchangeability when a shared transition core is richly encoded. 
In contrast, stitching involving MAE is environment-dependent: while MAE participates in largely successful swaps in \textit{PointMaze}, the reverse directions with MAE as the front-end degrade sharply in \textit{Wall}. 
We interpret this asymmetry as primarily reflecting endpoint competence in \textit{Wall} (MAE's unstitched planning success is lower there, e.g., $0.66$ vs.\ $0.90$ in \textit{PointMaze}$)$: a lightweight MLP can align feature spaces but cannot supply missing planning-relevant transition factors, and multi-step autoregressive rollouts compound the resulting mismatch.

Our evaluation is not based on one-step prediction. We use a CEM planner with horizon $H=5$ imagined rollout, where the world model generates latent trajectories autoregressively: each predicted latent is fed back as input for the next step. Achieving strong SSR under this closed-loop autoregressive setting constitutes a stricter functional test than one-step alignment and indicates that stitched representations remain compatible under multi-step compounding errors.

\section{Conclusion}

In this study, we investigated the \textbf{Platonic Representation Hypothesis} within the domain of world models, examining how predictive consistency can encourage heterogeneous vision priors to develop shared latent structure. Our findings yield three main insights:

\begin{itemize}
    \item \textbf{Predictive Consistency as an Alignment Driver:} Our results show that despite originating from disparate feature spaces, predictors trained with a shared state-transition objective can evolve geometrically similar local topologies. This convergence is most pronounced in transformer-based priors such as SigLIP and MAE, which exhibit a clear trajectory toward the designated Platonic anchor during training.
    
    \item \textbf{Architectural Constraints on Convergence:} While optimization drives alignment, the underlying model architecture constrains representational learning. The convergence observed among the ViT-based variants---and their collective divergence from the \textbf{ResNet} baseline---suggests that architectural compatibility facilitates recovery of a shared latent representation in our setting.
    
    \item \textbf{Functional Compatibility of Latent Spaces:} Through representation stitching experiments, we find that high-quality predictors can develop exchangeable transition factors. Recovering planning performance via lightweight transformations indicates that these aligned representations are not merely geometrically similar but can also be functionally compatible for downstream planning.
\end{itemize}

\paragraph{Limitations.}
Our conclusions are limited to frozen-encoder DINO-WM models and five simulated control tasks. The DINOv2-S anchor is an operational performance reference rather than a true Platonic ideal, and m-kNN/stitching probe latent geometry and functional compatibility rather than proving recovery of objective physical laws in full generality.

Predictive world-model training can therefore encourage heterogeneous visual priors to develop shared latent transition structure, providing a concrete setting for studying Platonic-like alignment in embodied prediction systems.

\clearpage
\section*{Impact Statement}
This paper presents work whose goal is to advance the field of Machine Learning. There are many potential societal consequences of our work, none of which we feel must be specifically highlighted here.

\bibliography{main}
\bibliographystyle{plainnat}

\clearpage
\onecolumn

\appendix

\section{Experiment Setting}
\label{app:experiment_setting}

\subsection{Predictor Architecture Details}
\label{predictor details}
Following the implementation in DINO-WM, our predictor is designed to process sequences of spatial patch features (e.g., from DINOv2) with an embedding dimension matching that of the vision encoder. The model architecture is a customized Vision Transformer with the following specifications:

\begin{itemize}
    \item \textbf{Backbone Hyperparameters:} The transformer has a depth of 6 layers, 16 attention heads, and an MLP hidden dimension of 2048. This configuration results in approximately 19 million trainable parameters when DINOv2-S is selected as the vision encoder (the embedding dimension is 384).
    \item \textbf{Action Conditioning:} To integrate control signals, the $K$-dimensional action vector is first projected to a high-dimensional space using a separate MLP. This projected action feature is then concatenated to every visual patch embedding at the corresponding time step.
    \item \textbf{Attention Mechanism:} We utilize a causal attention mask that enforces temporal causality. Specifically, for predicting the state at time $t$, each patch token can attend to all patches from previous time steps $t-H:t-1$. Crucially, within the \textit{same} time step, patches are fully visible to each other (bidirectional attention), allowing the model to reason about the global spatial structure of the predicted frame.
\end{itemize}

\subsection{Planning Details}
\label{app:planning_details}
Our planning implementation follows the Model Predictive Control (MPC) scheme described in \citet{Zhou2024DINOWMWM}. The core of the planner is the Cross-Entropy Method (CEM), which optimizes a sequence of actions $a_{t:T-1}$ to minimize a planning cost $C$.

\textbf{Cost Function.} The cost is defined as the Mean Squared Error (MSE) between the final predicted latent state $\hat{z}_T$ and the goal latent state $z_g$:
\begin{equation}
    C = \|\hat{z}_T - z_g\|^2
\end{equation}
where $\hat{z}_t = \psi(\hat{z}_{t-1}, a_{t-1})$ and $z_g = \phi(o_g)$. All forward computations for planning occur entirely within the latent space, eliminating the need for expensive pixel reconstruction.

\textbf{MPC and CEM Procedure.} We utilize a population size of $N=50$ action sequences and $30$ optimization iterations per step. The planning horizon $T$ is environment-dependent, typically ranging from 5 to 25 steps depending on the task complexity. After the optimization process concludes, the first $k$ actions of the optimal sequence are executed in the environment, and the process repeats at the next time step with new visual observations. To ensure a standardized comparison, planning results are uniformly reported at the 10th step of the process.

\section{Metric and Stitching Details}
\label{app:metric_stitching_details}

\subsection{Mutual kNN alignment}
Let $M_A$ and $M_B$ be two world models trained with different vision priors. We extract paired internal predictor representations $H_A$ and $H_B$ from the same held-out trajectories. Crucially, $H$ refers to predictor hidden states rather than frozen vision-encoder outputs. For a predictor with $L$ layers, we compute m-kNN for each layer and report the layer average.

For a specific layer, let $x_i^A$ and $x_i^B$ denote the feature vectors for sample $i$. We identify the $k$ nearest neighbors $\mathcal{N}_k(x_i^A)$ in space $A$ and $\mathcal{N}_k(x_i^B)$ in space $B$, and compute
\begin{equation}
\text{m-kNN}(A, B) = \frac{1}{N} \sum_{i=1}^N \frac{|\mathcal{N}_k(x_i^A) \cap \mathcal{N}_k(x_i^B)|}{k}.
\end{equation}
A higher score indicates stronger preservation of local topology, i.e., that two models place the same world states in more similar neighborhoods.

\subsection{Representation stitching protocol}
Although m-kNN quantifies local geometric alignment, it does not establish whether aligned representations contain transferable transition information. We therefore stitch predictor stacks from two trained world models. The stitched pipeline places early predictor layers from one model in front of later layers from another model. A lightweight two-layer MLP adapter $\mathcal{A}$ maps across the interface, and a two-layer projection $\mathcal{P}$ maps the stitched output back into the target feature space. All original world-model parameters remain frozen, and only the adapters are trained.

\begin{figure}[htbp]
    \centering
    \includegraphics[width=0.62\linewidth]{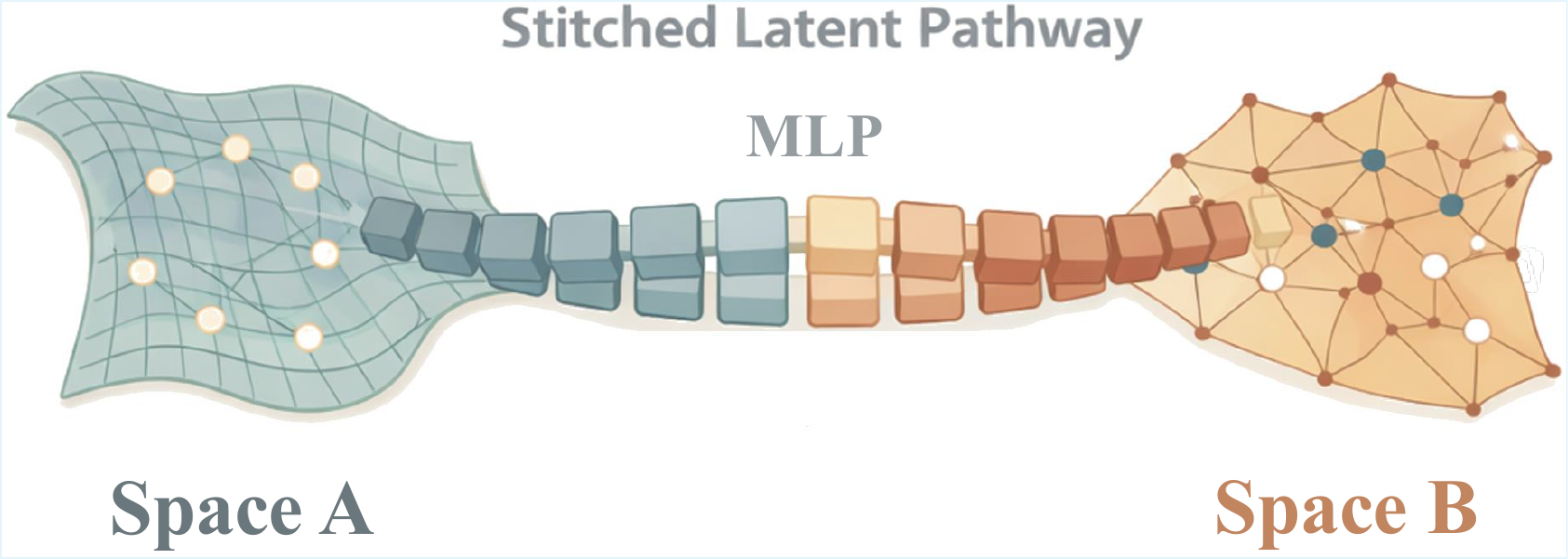}
    \caption{\textbf{Representation stitching.} We splice predictor stacks at depth $k$ and train only lightweight MLP adapters to map between feature spaces while freezing all other parameters.}
    \label{fig:stitching_overview}
\end{figure}

The pipeline can be summarized as
\begin{equation}
z \xrightarrow{M_{\text{front}}} \mathcal{A} \xrightarrow{M_{\text{back}}} \mathcal{P} \to z^*,
\label{eq:stitch_symbolic_final}
\end{equation}
where $z$ is the input visual feature and $z^*$ is the predicted future visual feature. We evaluate the stitched model with downstream planning success (SSR). High SSR under lightweight adaptation indicates that the two predictors expose exchangeable planning-relevant transition factors. For directionality tests, we swap front-end and back-end order while keeping the same adapter/projection design.

\revsection{Anchor Sensitivity of the Alignment Pattern}
\label{app:anchor_sensitivity}

\begin{revblock}
The main analysis uses the DINOv2-S world model at epoch 10 as the reference anchor. Since this anchor is selected for its strong planning performance, we test whether the observed alignment pattern is merely an artifact of this particular reference. We repeat the anchored m-kNN analysis using two additional high-performing ViT-based references, \textbf{SigLIP@10} and \textbf{MAE@10}, across all five environments. To avoid trivial values, we remove self-anchor comparisons, i.e., SigLIP compared with SigLIP@10 and MAE compared with MAE@10.
\end{revblock}

\begin{figure}[htbp]
  \centering
  \revfig{\includegraphics[width=\linewidth]{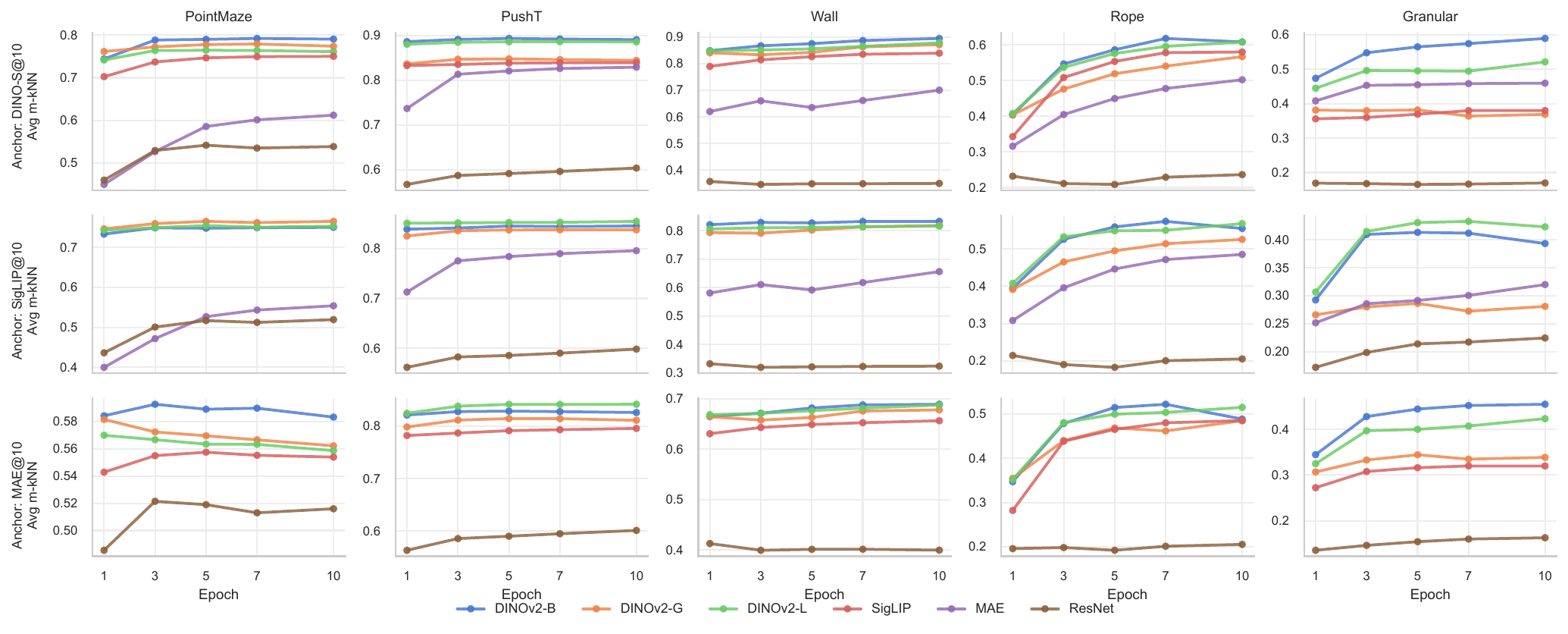}}
  \caption{\rev{\textbf{Anchor sensitivity analysis.}
  We recompute anchored m-kNN trajectories using DINOv2-S@10, SigLIP@10, and MAE@10 as reference anchors, excluding trivial self-anchor trajectories.
  While absolute scores and some local rankings vary with the anchor, the qualitative pattern remains stable: strong ViT-based predictors stay closer to one another, whereas ResNet is consistently less aligned in most environments.
  This suggests that the alignment trend is not unique to the DINOv2-S reference, although the anchor should still be interpreted as a performance-motivated reference rather than a true Platonic ideal.}}
  \label{fig:anchor_sensitivity}
\end{figure}

\begin{revblock}
As shown in Figure~\ref{fig:anchor_sensitivity}, changing the anchor affects the absolute m-kNN scale and can alter local rankings among strong ViT priors. However, the broader structure is preserved: DINOv2 variants, SigLIP, and MAE remain substantially more aligned with one another than the ResNet baseline. Thus, our use of DINOv2-S as the main anchor should be understood operationally, as a performance-motivated reference point for measuring alignment, not as an assumption that DINOv2-S represents a true Platonic ideal.
\end{revblock}

\revsection{Persistence under Longer Training}
\label{app:extended_training}

\begin{revblock}
We also test whether the epoch-10 alignment pattern is a fragile early-training artifact. We extend the analysis to later checkpoints around epochs 20, 30, and 50 for the four core priors where such checkpoints are available. For PushT, two late runs terminate at epoch 49, so we use the latest available checkpoint as the late-training proxy.
\end{revblock}

\begin{figure}[htbp]
  \centering
  \revfig{\includegraphics[width=0.86\linewidth]{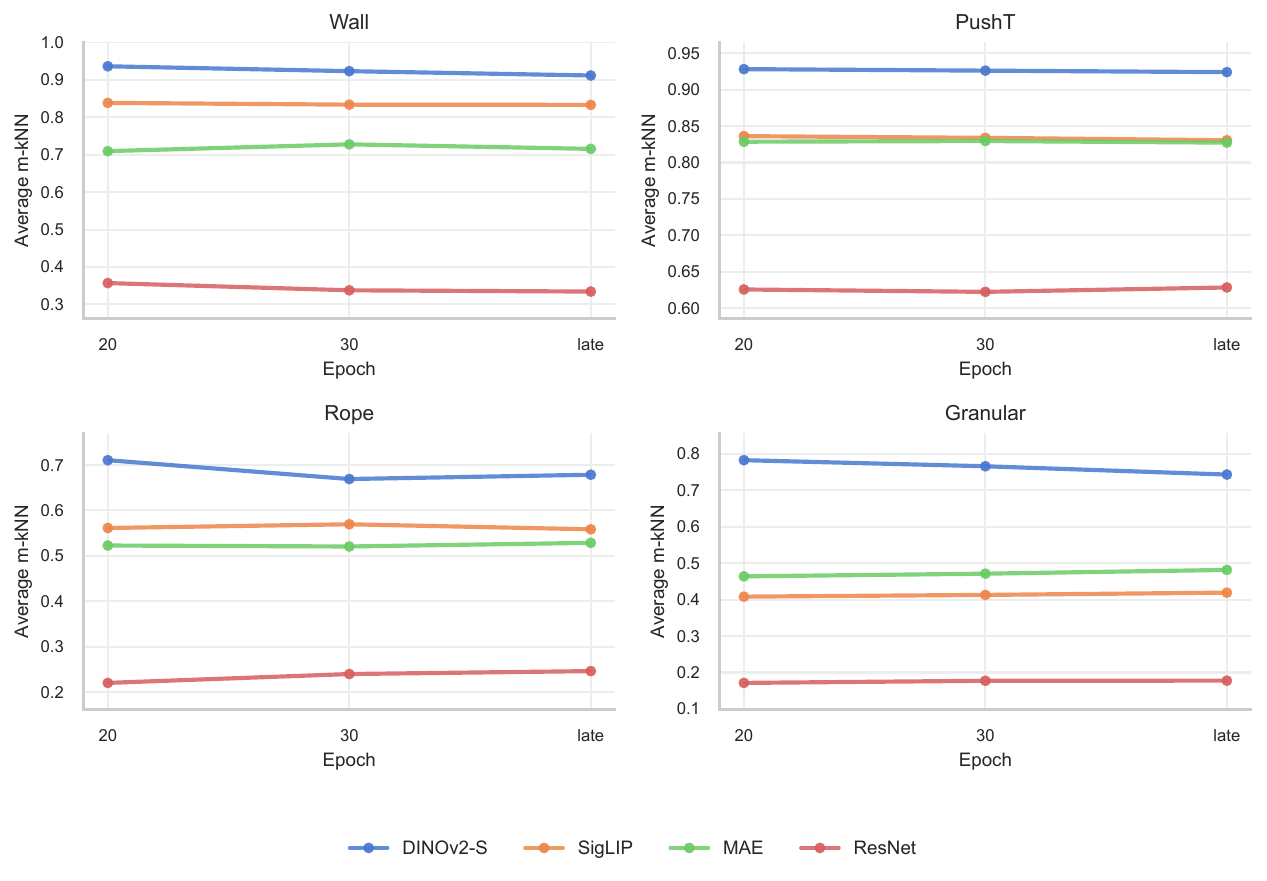}}
  \caption{\rev{\textbf{Longer-training stability of anchored m-kNN.}
  We evaluate DINOv2-S, SigLIP, MAE, and ResNet at later checkpoints using DINOv2-S@10 as the reference.
  Later training does not uniformly increase alignment, but the broad relative structure remains stable: DINOv2-S stays highest, SigLIP and MAE form a middle tier, and ResNet remains least aligned in Wall, Rope, and Granular.
  The x-axis label ``late'' denotes epoch 50 for most runs and epoch 49 for two PushT runs.}}
  \label{fig:extended_training}
\end{figure}

\begin{revblock}
The results in Figure~\ref{fig:extended_training} support a stability interpretation. The main cross-model ordering observed at epoch 10 remains visible at substantially later checkpoints, indicating that our conclusions are not solely a consequence of the original evaluation window. However, the curves do not support a monotonic convergence story: alignment can slightly increase, decrease, or plateau depending on the model and environment.
\end{revblock}

\section{Spectral kNN Embedding Overlays of Predictor Features}
\label{app:knn_overlay}

To complement our quantitative planning results, we provide a set of 2D visualizations to examine whether predictor
feature geometry from different encoders gradually converges during training, and whether this convergence aligns with
planning behavior.

\paragraph{Method and setup.}
We use the predictor layer3 features of DINO at epoch 10 as a fixed reference coordinate system.
For each task and each epoch, we load the same set of layer3 features from the corresponding feature files
(1000 samples per file). We construct an internal kNN weighted graph within each representation using cosine
similarity with $K=10$, compute a 2D Laplacian Eigenmap embedding, and then apply a Procrustes alignment so that
embeddings from different encoders and epochs are placed into the same reference frame. This yields an overlay plot in
which differences in neighborhood structure and overall geometry appear as changes in overlap, spread, and manifold
shape.

\begin{figure}[htbp]
  \centering
  \includegraphics[width=\linewidth]{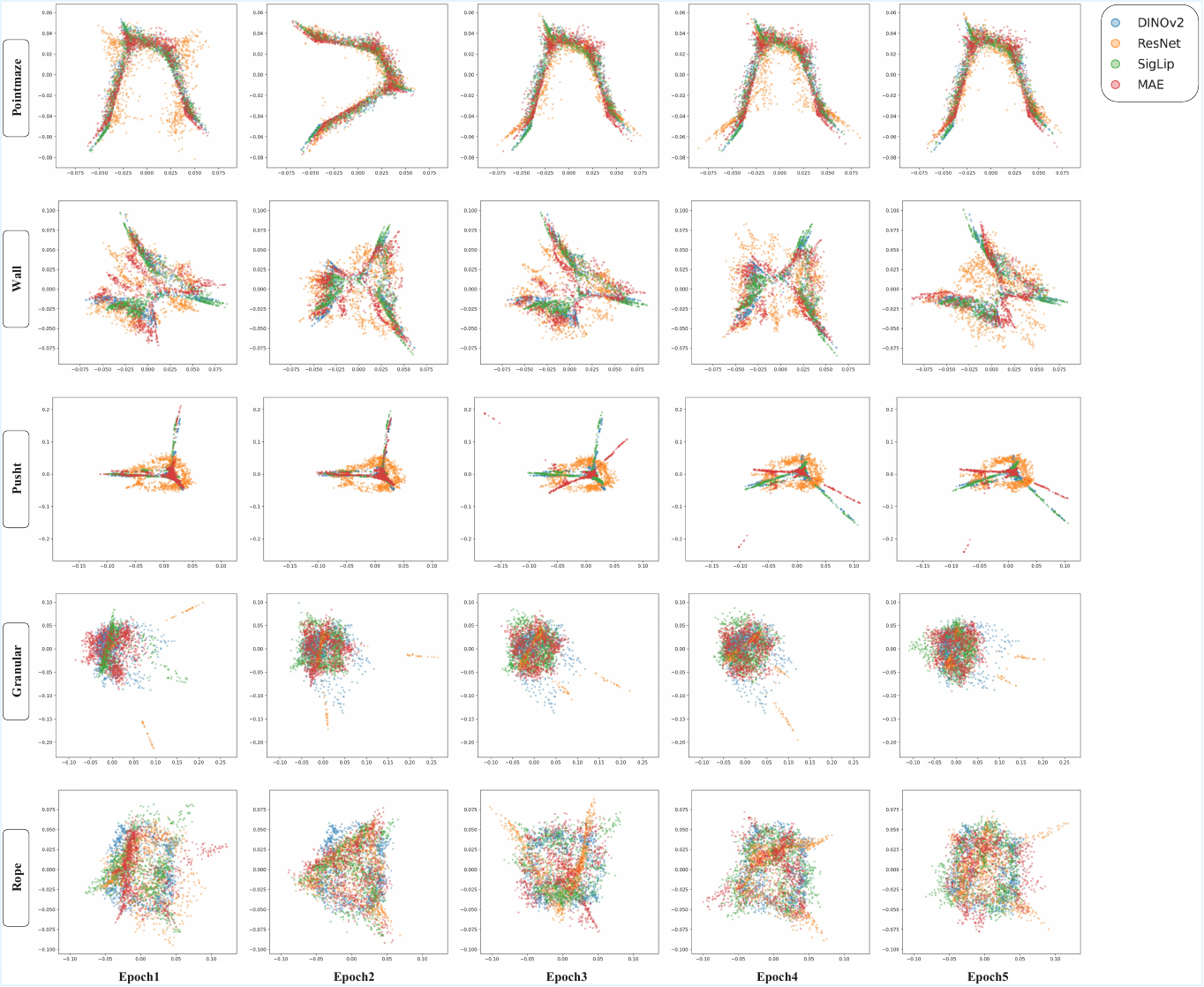}
  \caption{\textbf{kNN flow-matching visualization of predictor layer3 features across epochs.}
  We use the DINO predictor at epoch 10 as the anchor and visualize kNN correspondences in a 2D embedding.
  Across epochs, several feature distributions become more overlapping with the DINO anchor, while the trajectories
  remain task- and model-dependent; predictors with stronger planning performance often exhibit more similar distributions.}
  \label{fig:knn_flow_epoch}
\end{figure}

\paragraph{What we observe.}
As shown in Figure~\ref{fig:knn_flow_epoch}, DINOv2, SigLIP, and MAE exhibit a relatively clear geometric similarity in PointMaze, Wall, and
Push-T, as their aligned 2D spectral embeddings largely share similar manifold shape and neighborhood structure across
epochs. The ResNet prior behaves in a task-dependent manner. In PointMaze, where ResNet attains competitive planning
performance in our experiments, its distribution also shows a visible trend toward the transformer-like pattern.
In contrast, in Wall and Push-T, where ResNet performs worse, its embedding remains noticeably different from the three
transformer-based priors, with more dispersed structure and less overlap. Granular presents a more complex geometry,
appearing as clustered aggregates with noisier convergence across models. Rope shows the opposite tendency, with point
clouds that are comparatively more spread out, suggesting higher variability in the learned predictor geometry.
Overall, these plots serve as an auxiliary diagnostic that is consistent with our quantitative results, rather than a
standalone claim of representational equivalence.

\paragraph{Relation to Platonic representations and stitching.}
Under our Platonic-representation hypothesis, strong planning depends on a shared core of transition-critical factors
that is functionally reusable across models. Such reuse requires compatible neighborhood and metric structure so that
latent Euclidean distances remain meaningful for goal-conditioned ranking. The visual convergence in PointMaze and Wall
is consistent with our stitching results in these environments: when predictor geometries are more compatible, a
lightweight adapter is more likely to recover planning performance, yielding higher stitched success rate (SSR).

\section{Representation Geometry and Planning-Objective Mismatch}
\label{app:geom_mismatch}

During model training, we observed a consistent loss--planning decoupling in our experiments: representations trained with DINO yield
substantially better CEM planning performance than MAE, despite having an evaluation loss that is
roughly two orders of magnitude larger.
This appendix provides a principled explanation. Our CEM planner optimizes a latent-space MSE objective under a learned
world model, and planning quality is governed by (i) world-model rollout error and (ii) distortion of the latent
representation geometry. Reconstruction fidelity matters, but planning additionally requires that transition-critical factors are organized
so that latent Euclidean distances reflect task-relevant state discrepancies; otherwise CEM mis-ranks candidates.

\subsection{CEM objective in latent observation space}
\label{app:cem_obj}

For a candidate action sequence $\pi$ sampled by CEM, the world model predicts the final-step latent observation at horizon $T$ with two modalities,
\begin{equation}
z_T^{v,\mathrm{pred}}(\pi)\in\mathbb{R}^{d_v},\qquad
z_T^{p,\mathrm{pred}}(\pi)\in\mathbb{R}^{d_p}.
\end{equation}
Given a latent goal observation $(z_T^{v,*}, z_T^{p,*})$, the planner optimizes a per-dimension averaged MSE objective:
\begin{equation}
\hat J^{\mathrm{pred}}(\pi)
:=
\underbrace{\frac{1}{d_v}\left\|z_T^{v,\mathrm{pred}}(\pi)-z_T^{v,*}\right\|_2^2}_{\hat J_v^{\mathrm{pred}}(\pi)}
+
\alpha\underbrace{\frac{1}{d_p}\left\|z_T^{p,\mathrm{pred}}(\pi)-z_T^{p,*}\right\|_2^2}_{\hat J_p^{\mathrm{pred}}(\pi)}.
\label{eq:cem_latent_mse}
\end{equation}
CEM selects action sequences that minimize $\hat J^{\mathrm{pred}}(\pi)$.

\subsection{Rollout error induces bounded planning-objective error}
\label{app:rollout_error}

Let $(z_T^{v,\mathrm{real}}(\pi), z_T^{p,\mathrm{real}}(\pi))$ denote the final-step latent observations obtained under the same action sequence $\pi$ when evaluated under true dynamics. Define rollout errors:
\begin{equation}
e_v(\pi)\triangleq \left\|z_T^{v,\mathrm{pred}}(\pi)-z_T^{v,\mathrm{real}}(\pi)\right\|_2,\qquad
e_p(\pi)\triangleq \left\|z_T^{p,\mathrm{pred}}(\pi)-z_T^{p,\mathrm{real}}(\pi)\right\|_2.
\end{equation}
Assume bounded latent norms.

\paragraph{Assumption 1: Bounded latent observations.}
There exist constants $R_v,R_p>0$ such that for all $\pi$,
\begin{equation}
\left\|z_T^{v,\mathrm{pred}}(\pi)\right\|_2,\ \left\|z_T^{v,\mathrm{real}}(\pi)\right\|_2,\ \left\|z_T^{v,*}\right\|_2 \le R_v,
\qquad
\left\|z_T^{p,\mathrm{pred}}(\pi)\right\|_2,\ \left\|z_T^{p,\mathrm{real}}(\pi)\right\|_2,\ \left\|z_T^{p,*}\right\|_2 \le R_p.
\end{equation}

Define the ``real-latent'' objective
\begin{equation}
\hat J^{\mathrm{real}}(\pi)
:=
\frac{1}{d_v}\left\|z_T^{v,\mathrm{real}}(\pi)-z_T^{v,*}\right\|_2^2
+
\alpha\frac{1}{d_p}\left\|z_T^{p,\mathrm{real}}(\pi)-z_T^{p,*}\right\|_2^2.
\label{eq:real_latent_obj}
\end{equation}

\paragraph{Lemma 1: Objective mismatch from rollout error.}
Under Assumption 1, for any $\pi$,
\begin{equation}
\left|\hat J^{\mathrm{pred}}(\pi)-\hat J^{\mathrm{real}}(\pi)\right|
\le
\frac{4R_v}{d_v}e_v(\pi)
+
\alpha\frac{4R_p}{d_p}e_p(\pi).
\label{eq:rollout_bound}
\end{equation}

\paragraph{Proof.}
For any vectors $a,b,c$,
$\|a-b\|_2^2-\|c-b\|_2^2=(a-c)^\top(a+c-2b)$, hence
\begin{equation}
\left|\|a-b\|_2^2-\|c-b\|_2^2\right|
\le \|a-c\|_2\cdot\|a+c-2b\|_2.
\end{equation}
Apply this with $(a,c,b)=(z_T^{(\cdot),\mathrm{pred}}, z_T^{(\cdot),\mathrm{real}}, z_T^{(\cdot),*})$,
divide by dimension, and use Assumption 1 plus triangle inequality to bound
$\|z_T^{\mathrm{pred}}+z_T^{\mathrm{real}}-2z_T^*\|_2\le 4R$. \hfill$\square$

\subsection{Representation geometry controls whether latent MSE matches task distance}
\label{app:geometry}

The CEM objective \eqref{eq:cem_latent_mse} implicitly assumes that Euclidean distance in latent space is a meaningful proxy for task-relevant discrepancy.
Let $o_T(\pi)$ and $q_T(\pi)$ denote the true final visual observation and proprioceptive state produced by executing $\pi$ in the environment, and $(o_T^*,q_T^*)$ the goal.
Let $d_v(\cdot,\cdot)$ and $d_p(\cdot,\cdot)$ be task-relevant distances on the two modalities. Define the task objective:
\begin{equation}
J(\pi)=d_v\big(o_T(\pi),o_T^*\big)^2+\alpha\, d_p\big(q_T(\pi),q_T^*\big)^2.
\label{eq:true_task_obj}
\end{equation}

\paragraph{Assumption 2: Local bounded distortion.}
There exist constants $0<\underline\kappa_v\le \overline\kappa_v$ and $0<\underline\kappa_p\le \overline\kappa_p$ such that for all relevant pairs $(o,o')$ and $(q,q')$ encountered under planning,
\begin{align}
\underline\kappa_v\, d_v(o,o') &\le \|f_v(o)-f_v(o')\|_2 \le \overline\kappa_v\, d_v(o,o'), \nonumber\\
\underline\kappa_p\, d_p(q,q') &\le \|f_p(q)-f_p(q')\|_2 \le \overline\kappa_p\, d_p(q,q'),
\label{eq:bilip}
\end{align}
where $z^v=f_v(o)$ and $z^p=f_p(q)$.

\paragraph{Lemma 2: Latent MSE approximates task distance under bounded distortion.}
Under Assumption 2, for any $\pi$,
\begin{align}
\frac{\underline\kappa_v^2}{d_v}\, d_v\big(o_T(\pi),o_T^*\big)^2
\le
\frac{1}{d_v}\left\|z_T^{v,\mathrm{real}}(\pi)-z_T^{v,*}\right\|_2^2
\le
\frac{\overline\kappa_v^2}{d_v}\, d_v\big(o_T(\pi),o_T^*\big)^2,\\
\frac{\underline\kappa_p^2}{d_p}\, d_p\big(q_T(\pi),q_T^*\big)^2
\le
\frac{1}{d_p}\left\|z_T^{p,\mathrm{real}}(\pi)-z_T^{p,*}\right\|_2^2
\le
\frac{\overline\kappa_p^2}{d_p}\, d_p\big(q_T(\pi),q_T^*\big)^2.
\end{align}
\paragraph{Proof.}
Apply \eqref{eq:bilip} to $(o_T(\pi),o_T^*)$ and $(q_T(\pi),q_T^*)$ and square both sides. \hfill$\square$

\subsection{Decomposition: planning mismatch = rollout error + geometry distortion}
\label{app:decomposition}

Combining Lemma~1 and Lemma~2 yields the standard decomposition:
\begin{equation}
\left|\hat J^{\mathrm{pred}}(\pi)-J(\pi)\right|
\le
\underbrace{\left|\hat J^{\mathrm{pred}}(\pi)-\hat J^{\mathrm{real}}(\pi)\right|}_{\text{rollout error}}
+
\underbrace{\left|\hat J^{\mathrm{real}}(\pi)-J(\pi)\right|}_{\text{geometry distortion}}.
\label{eq:decomp}
\end{equation}
The first term is controlled by final-step latent prediction errors $e_v(\pi),e_p(\pi)$ via Lemma~1.
The second term is controlled by distortion factors $(\underline\kappa,\overline\kappa)$ via Lemma~2.

\paragraph{Why loss can decouple from planning.}
Eq.~\eqref{eq:decomp} shows that planning can fail not only due to rollout error but also due to geometry distortion.
Even with accurate rollouts, distorted latent geometry can make Euclidean distances a poor proxy for task relevant
discrepancy and thus degrade CEM ranking.

In our Platonic representation view, effective planning relies on a shared set of transition critical state factors.
For these factors to support goal conditioned ranking, they must be encoded in a way that preserves task relevant
neighborhood and distance structure. This makes geometry distortion a plausible contributor to the loss planning
decoupling we observe, where MAE can achieve lower reconstruction loss yet still plan worse than DINO.

\end{document}